\documentclass[preprint,12pt,authoryear]{elsarticle}

\usepackage{amssymb}
\usepackage{amsmath}
\usepackage{algorithmicx}
\usepackage{algorithm}
\usepackage{array}
\usepackage{url}
\usepackage{multirow}
\usepackage{booktabs}
\usepackage{algpseudocode}
\usepackage{hyperref}

\journal{Expert Systems with Applications}

\begin{document}

\begin{frontmatter}

\title{Uncertainty Aware Human-machine Collaboration in Camouflaged Object Detection} 

\author[label1]{Ziyue Yang}\ead{zyyang@hdu.edu.cn}
\author[label1]{Kehan Wang}\ead{242050376@hdu.edu.cn}
\author[label1]{Yuhang Ming}\ead{yuhang.ming@hdu.edu.cn}
\author[label1]{Yong Peng}\ead{yongpeng@hdu.edu.cn}
\author[label1]{Han Yang}\ead{43008@hdu.edu.cn}
\author[label2]{Qiong Chen}\ead{chenqiong@cethik.com}
\author[label1]{Wanzeng Kong\corref{cor1}}\ead{kongwanzeng@hdu.edu.cn}

\affiliation[label1]{organization={College of Computer Science, Hangzhou Dianzi University},
             addressline={1158 Baiyang Road}, 
             city={Hangzhou},
             postcode={310018}, 
             state={Zhejiang},
             country={China}}

\affiliation[label2]{organization={Hangzhou Hikvision Digital Technology Company},
             addressline={555 Qianmo Road}, 
             city={Hangzhou},
             postcode={310051}, 
             state={Zhejiang},
             country={China}}

\begin{abstract}
Camouflaged Object Detection (COD), the task of identifying objects concealed within their environments, has seen rapid growth due to its wide range of practical applications. A key step toward developing trustworthy COD systems is the estimation and effective utilization of uncertainty. In this work, we propose a human-machine collaboration framework for classifying the presence of camouflaged objects, leveraging the complementary strengths of computer vision (CV) models and noninvasive brain-computer interfaces (BCIs). Our approach introduces a multiview backbone to estimate uncertainty in CV model predictions, utilizes this uncertainty during training to improve efficiency, and defers low-confidence cases to human evaluation via RSVP-based BCIs during testing for more reliable decision-making. We evaluated the framework in the CAMO dataset, achieving state-of-the-art results with an average improvement of 4.56\% in balanced accuracy (BA) and 3.66\% in the F1 score compared to existing methods. For the best-performing participants, the improvements reached 7.6\% in BA and 6.66\% in the F1 score. Analysis of the training process revealed a strong correlation between our confidence measures and precision, while an ablation study confirmed the effectiveness of the proposed training policy and the human-machine collaboration strategy. In general, this work reduces human cognitive load, improves system reliability, and provides a strong foundation for advancements in real-world COD applications and human-computer interaction. Our code and data are available at: https://github.com/ziyuey/Uncertainty-aware-human-machine-collaboration-in-camouflaged-object-identification.
\end{abstract}

\begin{graphicalabstract}
\includegraphics[width=\textwidth]{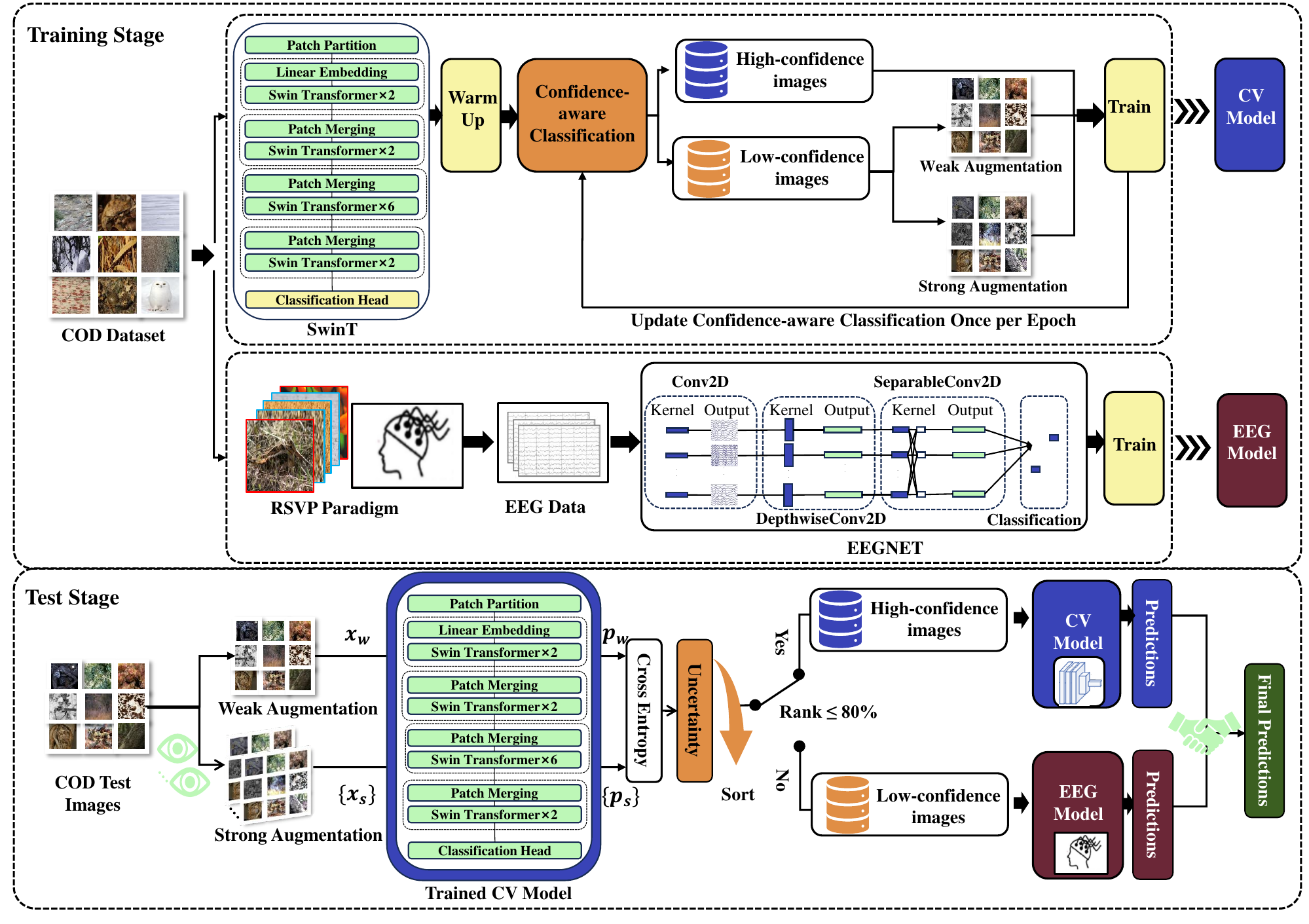}
\end{graphicalabstract}

\begin{highlights}
\item Introduces a novel framework combining computer vision (CV) models with RSVP-based brain-computer interfaces (BCIs) to enhance camouflaged object detection (COD).
\item Develops a multiview backbone that estimates prediction confidence, leveraging strong and weak augmentations to assess uncertainty and guide model training.
\item Incorporates RSVP-based BCIs to improve detection accuracy, where low-confidence cases from the CV model are re-evaluated using human cognitive responses.
\item Achieves 4.56$\%$ higher balanced accuracy (BA) and 3.66$\%$ higher F1 score on the CAMO dataset compared to existing methods, with the best cases reaching 95.60$\%$ BA and 95.49$\%$ F1 score.
\item Proposes a selective human intervention strategy, deferring only uncertain cases to human evaluation, significantly reducing cognitive effort while maintaining high reliability.
\end{highlights}


\cortext[cor1]{Corresponding author.}

\end{frontmatter}

\section{Introduction}
\label{sec1}
Rapid adoption of deep learning has highlighted concerns about the robustness and transparency of neural network predictions. Addressing these concerns is critical for the advancement of trustworthy artificial intelligence (AI) \cite{kaur2022trustworthy}. A key aspect of building trust is to enable models to assess and communicate their own uncertainty \cite{10.1145/3675392,gawlikowski2023survey}, which can guide decisions to defer to human operators or seek additional data in complex scenarios. In particular, in tasks where humans and AI have complementary strengths, effective human-machine collaboration represents a promising trend in the evolution of digital societies. However, how to leverage uncertainty quantification for such collaboration, and its impact, remains underexplored.
Camouflaged Object Detection (COD) is a well-suited task for such synergy. COD focuses on identifying objects concealed within their environments, making it both challenging and intriguing \cite{lv2023toward}. This field has grown rapidly due to its practical applications, such as defect detection in manufacturing \cite{xiong2021attention}, pest control in agriculture \cite{rustia2020application}, segmentation of lesions in medical diagnosis \cite{fan2020pranet}, pedestrian detection in nighttime environments\cite{yang2024dual}, and even creative pursuits such as image blending \cite{suo2021neuralhumanfvv}. Advances in image-level camouflaged object segmentation have been driven by models such as DPSNet \cite{li2024dpsnet}, JCNet \cite{jiang2023camouflaged}, SINet-V2 \cite{fan2021concealed}, and DGNet \cite{ji2023deep}, supported by recognized datasets and benchmarks \cite{bi2021rethinking}. COD tasks involve two key components: identifying whether camouflaged objects are present and segmenting them when they are. However, much of the current research focuses on the segmentation stage, assuming the model can produce a continuous mask map where a black saliency map signifies the absence of a camouflaged object. However, since camouflaged objects are not guaranteed to be present, a critical first step is to detect their presence before proceeding to segmentation. While precise localization is vital for applications like medical diagnostics, in tasks such as rescue operations or pest monitoring, detecting the presence of camouflaged objects takes precedence. Therefore, our study centers on binary classification to determine the presence or absence of camouflaged objects.

Despite rapid progress in COD, concerns persist about the safety and reliability of computer vision (CV) models. Deep learning models often function as black boxes \cite{hassija2024interpreting} and can struggle with challenges such as small targets, incomplete objects, and complex backgrounds (e.g. noise, obstructions, or shadows) \cite{bi2021rethinking}. Teaching models to admit uncertainty, essentially saying 'I don't know', remains a significant challenge.

In contrast, the human brain excels at adapting to diverse environments, recognizing subtle patterns, and identifying hidden objects in complex scenarios, such as low-light conditions or cluttered backgrounds. This adaptability allows humans to detect hidden or camouflaged targets with high accuracy, even when features are partially occluded. However, manual search for such targets is time-consuming. Non-invasive brain-computer interfaces (BCIs) offer a novel solution to this challenge \cite{kim2019high}. When individuals encounter rare targets in visual sequences, their brain activity generates event-related potentials (ERPs), particularly the P300 component, which can help detect targets. Researchers have exploited this phenomenon by combining rapid serial visual presentation (RSVP) paradigms with BCIs technology to evoke ERPs in response to visual stimuli, allowing efficient classification of target images \cite{zhang2020benchmark}--even under challenging conditions where targets are camouflaged and hidden \cite{lian2023eeg, zhou2024rsvp}. 

In this study, we propose a novel framework for human-machine collaboration in COD, leveraging model uncertainty as the bridge between CV models and human intervention. Specifically, CV models handle the bulk of images, exploiting their ability to process large volumes of data in parallel, while humans, through BCIs, provide instinctive responses to challenging or unusual images flagged by model uncertainty. To the best of our knowledge, no existing research has been proposed that uses and evaluates the uncertainty-based human-machine partnership in COD. Our contributions include the following.

\begin{itemize}
    \item Multiview backbone for COD: We propose a backbone that evaluates the confidence across multiple views for each image.
    \item Uncertainty-aware training policy: Our training strategy improves CV models by effectively using information from the training set.
    \item Human-machine collaboration paradigm for COD: For samples with low confidence from the CV model, we fuse RSVP-based BCIs results with model predictions to achieve more reliable decision making.
\end{itemize}

\section{Related Work}
\subsection{Uncertainty Quantification}
Uncertainty estimation, or confidence assessment, in neural network predictions has emerged as a major research focus in the machine learning community \cite{smith2024uncertainty}. Current approaches to uncertainty modeling typically fall into three categories: Monte Carlo Dropout \cite{neal2012bayesian, Moreau_2022_WACV, gal2017concrete, kang2023active}, the Bootstrap model \cite{osband2016deep}, and the Gaussian Mixture Model \cite{9666964, zhang2019short}. These methods have been extensively explored in various domains, demonstrating promising results \cite{abdar2021review}. However, most models merely display uncertainty without leveraging it to guide further actions. To address this gap, recent advances have begun to incorporate uncertainty into training strategies \cite{li2023disc,cordeiro2023longremix}. Although promising, these cutting-edge efforts have focused primarily on tasks that involve noisy label learning. For standard supervised learning tasks, the effectiveness of uncertainty-informed training strategies remains unclear. In this work, we extend the confidence learning method based on two views introduced in \cite{li2023disc} and apply it to the COD problem. Our approach not only integrates uncertainty into the model's learning process, but also flexibly delegates uncertain samples to human-based RSVP systems for enhanced decision making.

\subsection{Camouflaged Object Detection (COD)}
In recent years, numerous deep learning-based COD models have been developed \cite{liang2024systematic}, while recent research has started to place more emphasis on uncertainty. Yi Zhang et al. introduced PUENet \cite{10159663}, which uses a Bayesian conditional variational auto-encoder for predictive uncertainty estimation. Yixuan Lyu et al. \cite{10183371} proposed the Uncertainty-Edge Dual Guide model, which combines probabilistic uncertainty with deterministic edge information for accurate COD. Jiawei Liu et al. \cite{9706783} developed a confidence-based COD framework with dynamic supervision, producing both camouflage masks and aleatoric uncertainty estimates, showing superior performance. Fan Yang et al. \cite{9710683} integrated Bayesian learning with Transformer reasoning, leveraging both deterministic and probabilistic information to improve detection accuracy. Furthermore, Aixuan Li et al. \cite{9578707} proposed an adversarial learning network for higher-order similarity measures and confidence estimation. Current research mainly addresses uncertainty in segmentation tasks, focusing on generating confidence maps for boundary distinction, while this study aims to model the uncertainty in identifying object presence to enhance classification accuracy.

\subsection{RSVP-based BCIs for Target Detection}
RSVP-based BCIs have received significant attention in recent years, particularly in the domain of target detection, due to their efficiency in processing rapid visual stimuli and eliciting robust neural responses such as potentials related to P300 events. Research advancements have focused on optimizing RSVP paradigms to enhance system performance, with notable contributions including the use of adaptive parameter tuning and hybrid paradigms that integrate steady-state visual evoked potentials to improve detection accuracy and user experience \cite{jalilpour2020novel}. Novel electroencephalogram(EEG) decoding algorithms, such as deep learning frameworks that take advantage of convolutional neural networks \cite{santamaria2020eeg} and attention mechanisms\cite{wang2020linking}, have further enhanced classification performance, while collaborative approaches of multiple users have demonstrated the potential for improved accuracy through collective neural signal analysis \cite{9931160}. The introduction of benchmark data sets has standardized the evaluation of algorithms and facilitated reproducible research \cite{zhang2020benchmark}. Furthermore, the integration of multimodal data, including EEG and eye tracking, has shown promise in addressing signal noise and enhancing target detection reliability \cite{mao2023cross}, cementing RSVP-BCIs as a crucial interface for bridging neuroscience and real-world applications. The most related work to ours is by Yujie Cui et al.\cite{cui2022dynamic}, who proposed a human-computer fusion method called Dynamic Probability Integration for nighttime vehicle detection. Their approach uses a probability assignment method to assign classification weights between different information sources, which requires full human participation in the EEG-based RSVP task. In contrast, our model reduces human effort by using uncertainty to guide collaboration, with humans only evaluating high-uncertainty samples from the CV model, thus improving efficiency.

\section{Method}
\begin{figure*}[ht]
    \centering
    \includegraphics[width=\textwidth]{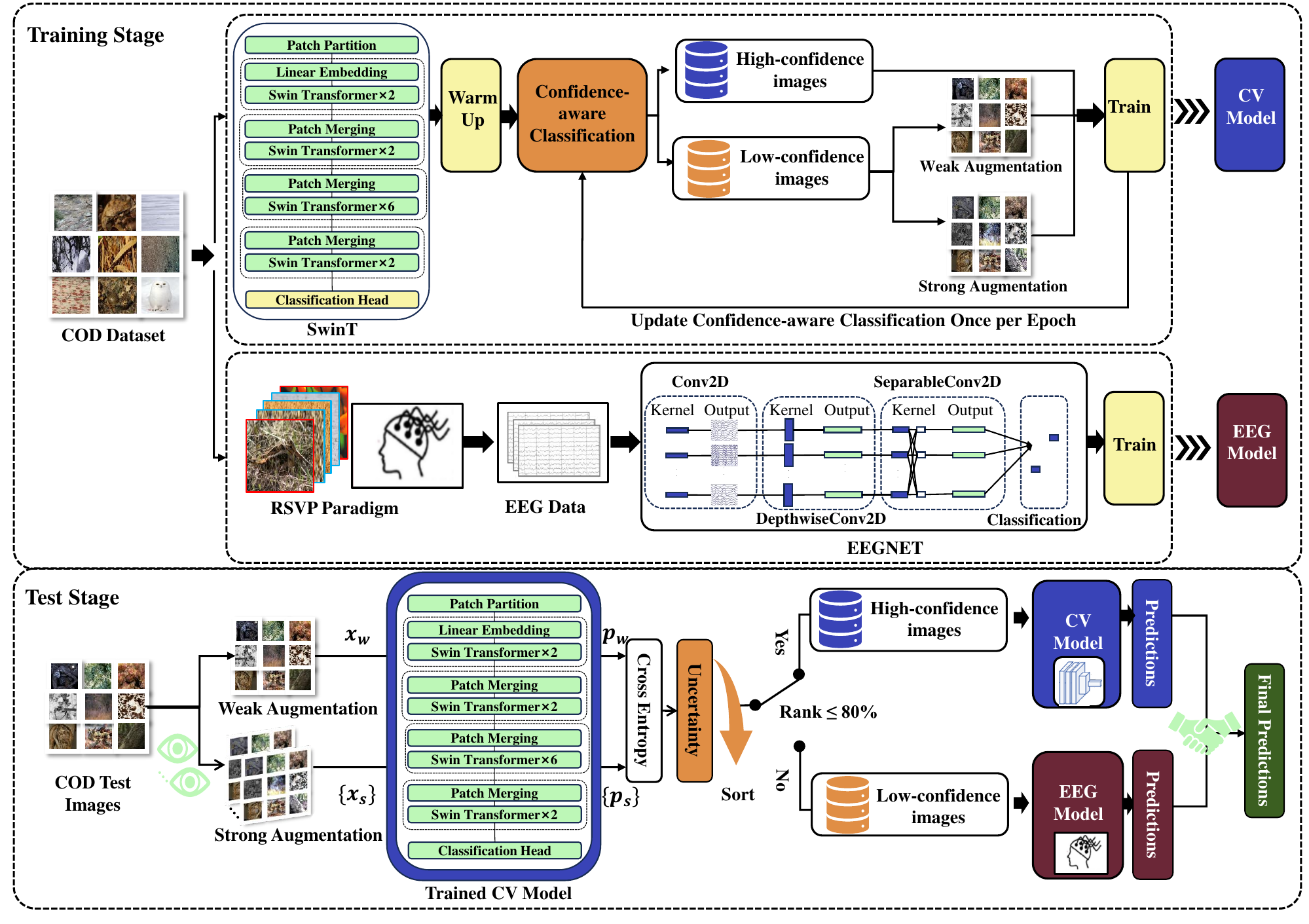} 
    \caption{Overall framework of this study. During training, the dataset is split into high and low-confidence sets, with augmentations applied to low-confidence samples and the split updated per epoch; during testing, high-uncertainty samples are classified using the RSVP program, while high-confidence samples are handled by the CV model, enhancing COD performance.}
    \label{fig:overall}
\end{figure*}
The framework of this paper is illustrated in Figure
~\ref{fig:overall}. In the training stage, the CV model is pretrained on the complete dataset. The data is then divided into a high confidence set (\(X_{hc}\)) and a low confidence set (\(X_{lc}\)) based on uncertainty, the split being updated once per epoch. Weak and strong augmentations are applied to \(X_{lc}\), creating \(X_{lc'}\), while \(X_{hc}\) is oversampled to match the size of \(X_{lc'}\). In the testing phase, high-uncertainty samples are transferred to the EEG-based RSVP program for classification, while high-confidence samples are predicted using the CV model. This two-stage process combines the strengths of CV models and RSVP-based BCIs systems to enhance COD performance.

\subsection{Dataset}
The data set used in this study consists of 2,500 images, evenly divided into camouflage target images and background images. Our data set is sourced from the publicly available CAMO dataset, which includes 1,250 camouflage target images. Using the ground truth map for each image, we generated paired camouflage target and background images through a combination of manual and automated methods. The resulting data set is available on the provided GitHub link. Although the CAMO-COCO dataset was considered, its background images differ substantially from the camouflage target images, making it less aligned with our focus. This study emphasizes scenarios where camouflage targets are embedded within similar backgrounds, reflecting more realistic application contexts.

\subsection{CV Model}
\subsubsection{Multi-view based Uncertainty Estimation}
We propose a backbone model that evaluates confidence in multiple views for each image. For each sample \((x, y)\), we apply one weak augmentation to \(x\), resulting in \(x_w\), and \(n\) strong augmentations, resulting in \(\{x_{s1}, x_{s2}, \dots, x_{sn}\}\). To compute uncertainty, we used the cross-entropy (CE) between two distributions \(p\) and \(q\) defined as:
\begin{equation}
\text{CE}(p, q) = -\sum_{i=1}^{C} p_i \log(q_i),
\label{eq:ce}
\end{equation}
where \(C\) is the number of classes, \(p_i\) is the ground truth probability of class \(i\), and \(q_i\) is the predicted probability of class \(i\). Hence, uncertainty is calculated as the mean cross-entropy between the weakly augmented sample and each strongly augmented sample:
\begin{equation}
\text{Uncertainty} = \frac{1}{n} \sum_{j=1}^{n} \text{CE}(p_w, p_{sj}),
\label{eq:uncertainty}
\end{equation}
where $n$ denotes the number of strong augmentations, while $x_w$ and $x_{sj}$ are processed by the trained CV model to obtain $p_w$ and $p_{sj}$, respectively. This approach quantifies uncertainty by assessing the consistency between weakly and strongly augmented views of the same sample.

\subsubsection{Strong and weak augmentation}
In this study, strong and weak data augmentations serve two primary purposes: evaluating multiview-based uncertainty and enhancing the model's robustness during training.

Weak augmentations include operations such as random horizontal flipping, slight rotation, and random cropping. In contrast, strong augmentations involve more significant perturbations, such as cropping, color transformations, image quality adjustments, occlusion, and composite augmentations to simulate complex scene variations. The specific augmentation method for both weak and strong augmentations is selected randomly for each iteration.

\begin{algorithm}[H]
\caption{Uncertainty-Aware Training Policy}
\label{alg:training_procedure}
\begin{algorithmic}[0]
    \State \textbf{Input:} Dataset $W$, Total Epochs $E$, Batch Size $B$, Warm-up Epochs $\text{Warm}$, Ramp-up Length $\text{rampup\_length}$, Augmentation Strength $M$, Augmentation Times $N$, Clean Dataset Filtering Method $\text{Method\_c}$, Consecutive Clean Rounds $t$

    \State Load pre-trained weights from ImageNet
    \State Initialize epoch counter $e \gets 0$
    
    \While{$e < E$}
    
        \For{$i = 1$ \textbf{to} $|W|$}
            \State Perform warm-up and compute loss using cross-entropy loss: $l(x, y) = \text{CE}(f(x), y)$
        \EndFor
    
        \If{$e > \text{Warm}$}
            \State Build high-confidence $X_{hc}(K)$ and low-confidence $X_{lc}(K)$ sets
            \State $X_{hc}(K) \gets \text{Samples from last } t \text{ epochs of training}$
            \State $X_{lc}(K) \gets \text{All training samples} \setminus X_{hc}(K)$
            \State $num\_iter \gets \frac{|X_{hc}(K)|}{B}$
    
            \For{$iter = 1$ \textbf{to} $num\_iter$}
                \State Select high-confidence samples $\{(X, Y)\} \sim X(k)$
                \State Select low-confidence samples $\{(U, Y)\} \sim U(k)$
    
                \For{$b = 1$ \textbf{to} $B$}
                    \State $x_b  \gets \text{Batch samples for high-confidence set}$
                    \State $u_{b1} \gets  \text{StrongDataAugment}(u_b)$
                    \State $u_{b2} \gets  \text{WeakDataAugment}(u_b)$
                    \State $r \gets \text{clip}\left(\frac{e - \text{Warm}}{\text{rampup\_length}}, 0, 1\right) \times \lambda_u$
                    \State $loss = l(x_b, y_b) + [l(u_{b1}, y_b) + l(u_{b2}, y_b)] \times r$
                \EndFor
            \EndFor
        \EndIf
        
        \State Update high- and low-confidence sets using \texttt{Method\_c}
        \State Increment epoch: $e = e + 1$
    \EndWhile
    
    \State \textbf{Output:} Model Parameters $\theta$
\end{algorithmic}
\end{algorithm}

The uncertainty-aware training policy outlined in Algorithm 1 aims to enhance model robustness by leveraging high-confidence and low-confidence samples during training. Initially, during the warm-up phase, the model is trained using standard loss functions. Once the warm-up is complete, the training data is divided into high-confidence and low-confidence sets. High-confidence samples \(X_{hc}(K)\) are selected from the last training epochs $t$, while low-confidence samples \(X_{lc}(K)\) are derived from the remaining data. The model is then trained on these two sets, with low-confidence samples undergoing strong and weak data augmentations (detailed in Section 3.2.2 ) to improve generalization. For confident set selection, \texttt{Method\_c} includes five strategies based on uncertainty estimation: (1) splitting Low-confidence and High-confidence images in a 1:2 ratio (Ratio 1:2); (2) splitting in a 2:1 ratio (Ratio 2:1); (3) dynamically partitioning the samples using a threshold (details: dividing data into intervals using a 0.1 threshold, then traversing each interval in order of increasing accuracy, selecting the first interval with a sample and accuracy lower than the overall accuracy, and using the median of that interval as the dynamic threshold) (Dynamic Threshold); (4) labeling images with consistent predicted and ground-truth labels after augmentation as High confidence, otherwise labeling them as Low-confidence (Consistent Labeling); and (5) labeling images as High confidence if at least one prediction matches the ground truth after augmentation, otherwise labeling them as Low confidence (At Least One Match). A key feature of the policy is the gradual increase in weight loss for low-confidence samples, controlled by the factor $r$, which evolves throughout training. This dynamic weighting helps stabilize training by initially focusing on high-confidence samples and progressively incorporating low-confidence data. The subsets of high- and low-confidence samples are updated after each epoch using a filtering method to ensure data quality. This strategy effectively balances the influence of clean, high-confidence data with more uncertain, low-confidence data, leading to improved model performance and robustness.

\subsection{RSVP-based EEG Model}\label{stage-2-brain-machine-synergy}

\subsubsection{Participants and Data
Recording}
The study was reviewed and approved by the 	
Ethics Committee of Second Affiliated Hospital of Zhejiang University, College of Medicine and the protocol number is IRB-2024-1535. Signed informed consent was obtained from each participant.
A total of 8 participants (mean age: 24.10 years) were recruited for
the brain-machine collaborative RSVP target detection paradigm study.
All participants had normal or corrected vision and no neurological
problems. Before the experiment, each participant was informed of the
potential risks and signed a written informed consent form.\\
EEG data were recorded using the Synamps2 system (64 channels,
NeuroScan, Inc.) at a sampling rate of 1000 Hz. In this study, 62
electrodes were used to record EEG signals, following the
international 10-20 electrode placement system, with the reference
electrode placed at the vertex. Before data recording, the impedance
of the electrodes was measured and adjusted to ensure it remained below
25 k$\Omega$. EEG data were initially filtered using a finite impulse response filter with a frequency range between 0.1 and 40 Hz and
then resampled to 250 Hz for classification. Segments of events of
one second were extracted from the data starting from the onset of the
stimulus.

\subsubsection{RSVP Paradigm Design}
\begin{figure}[ht]
    \centering
    \includegraphics[width=\linewidth]{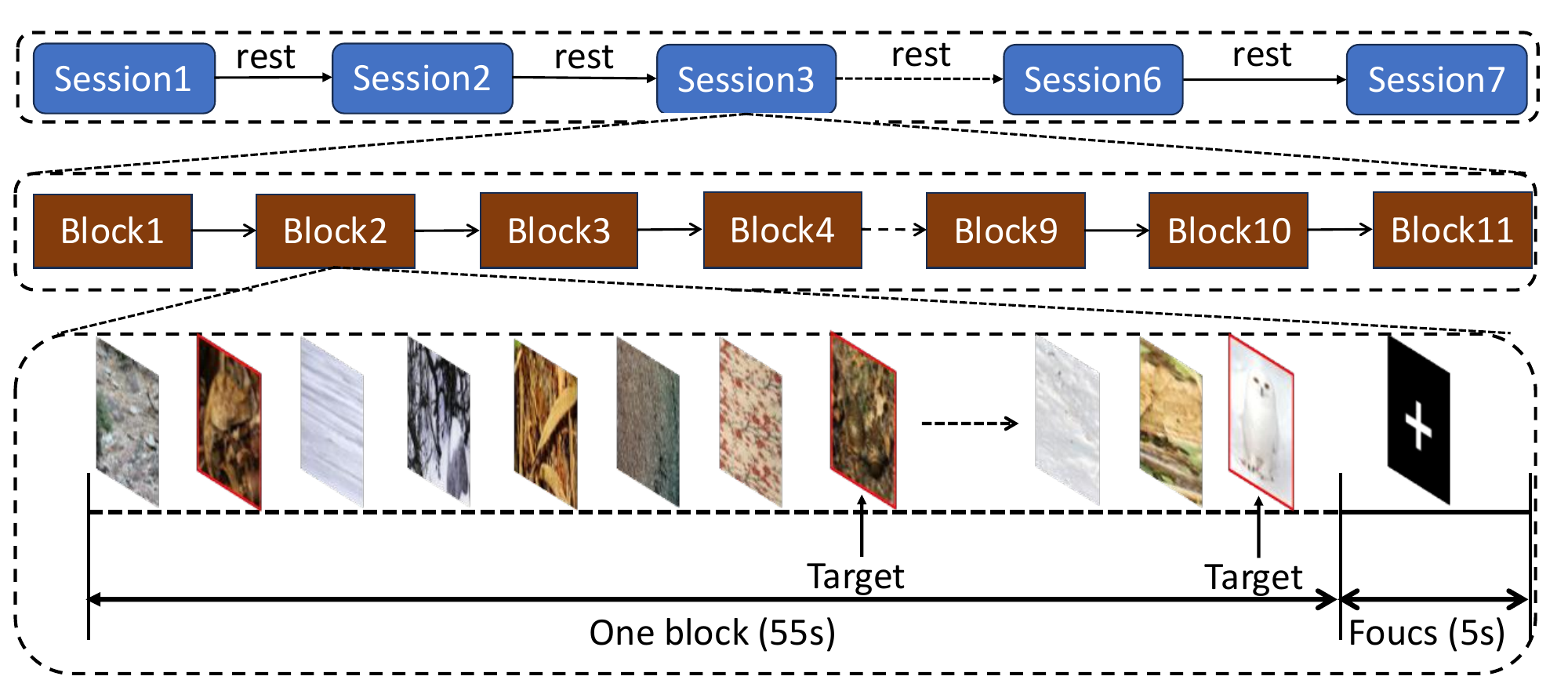} 
    \caption{RSVP Paradigm Design. At the start of each trial, a fixation
cross appeared in the center of the screen. The stimuli were then presented
at a frequency of 1 Hz. The participants were given ample rest time between
blocks. The experiment ensured that there were at least three nontarget
images between any two target images. All stimulus images were displayed
on a 512 × 512 resolution monitor with a refresh rate of 60 Hz.}
    \label{fig:RSVP}
\end{figure}
The experimental setup is shown in Figure~\ref{fig:RSVP}. In the experiment,
participants were seated in a quiet room and instructed to identify
camouflage targets within a sequence of stimuli. Original frames
were used to present the stimuli. An EEG amplifier was employed to
capture the participants\textquotesingle{} brain activity, with no
button pressing required when a target was detected.\\
The experimental procedure and paradigm parameters are depicted in the
figure. The experiment consisted of five blocks, each comprising 11 trials.
In each block, 75 target images and 530 untarget images were presented. Before the experiment began, a set
of images containing camouflage target and background images was
presented to the participants as a guideline, informing them of the
target they needed to identify. At the start of each trial, a fixation
cross appeared in the center of the screen. The stimuli were then presented
at a frequency of 1 Hz. The participants were given ample rest time between
blocks. The experiment ensured that there were at least three nontarget
images between any two target images. All stimulus images were displayed
on a 512 × 512 resolution monitor with a refresh rate of 60 Hz.

\subsubsection{Brain-Machine Collaboration}
In the test phase, samples with high uncertainty from the CV model are sent to human evaluation. These samples are treated as "potential targets." To construct the test target sequence, at least three nontarget images from the training set are inserted between each pair of target images, ensuring activation of the P300 component during stimulus presentation. An RSVP-based EEG model, trained in the training set, is then used to detect the presence of targets in the test set. The predictions for high-uncertainty samples are replaced with the output of the RSVP-based model, enabling effective human-machine collaboration.

\section{Results}
In this section, we evaluate the effectiveness of the proposed method. The data set was divided into training and testing sets with a 9: 1 ratio and the training set was further divided into training and validation subsets, also with a 9: 1 ratio. During the experiments, we used early stopping with a patience parameter set to 10. To ensure the stability and reliability of the results, we performed experiments using five different random seeds: 37, 12, 6, 99, and 123. We applied five strong augmentations and one weak augmentation. The reported results are the mean and standard deviation for the five runs. All experiments were performed on a GeForce RTX 4090 GPU. The evaluation of our experimental results was based on balanced accuracy (BA) and F1 score, with their respective formulas as follows:

\begin{equation}
\text{BA} = \frac{1}{2} \left( \frac{TP}{TP + FN} + \frac{TN}{TN + FP} \right),
\end{equation}
where \( TP \) is the number of true positives, \( FN \) is the number of false negatives, \( TN \) is the number of true negatives and \( FP \) is the number of false positives.

\begin{equation}
\text{F1} = 2 \times \frac{\text{Precision} \times \text{Recall}}{\text{Precision} + \text{Recall}},
\end{equation}
where \(\text{Precision} = \frac{TP}{TP + FP}\) and \(\text{Recall} = \frac{TP}{TP + FN}\).

\subsection{Results for different training policies}

In designing the training strategy, we compared different data augmentation and confident set selection methods, using the Swin Transformer (SwinT) as the backbone network. For data augmentation, we tested three strategies: (1) applying strong and weak augmentations separately to high-confidence and low-confidence images; (2) augmenting only high-confidence images (H-only); and (3) augmenting only low-confidence images (L-only). Details of the confident set selection approach are provided in Section 3.2.3. The results, shown in Table~\ref{2}, reveal that augmenting only low-confidence samples, along with a fixed ratio of 2:1 between low- and high-confidence images, yielded the best performance. This configuration achieved a BA of 89.92\% and an F1 score of 90.40\%. These findings suggest that it is crucial to effectively mine and utilize low-confidence samples during training. Based on these results, we selected the L-only augmentation strategy combined with a fixed ratio of 2:1 for subsequent experiments.

\begin{table}[ht]
\centering
\footnotesize
\caption{Comparison of Different Data Augmentation Strategies and Confident Set Selection Policies}
\label{1}
\begin{tabular}{clcc}
\toprule
\textbf{\parbox{2.5cm}{Augmentation}} & \textbf{\parbox{2.5cm}{Selection}} & \textbf{BA (\%)} $\uparrow$ & \textbf{F1 (\%)} $\uparrow$ \\
\midrule
Both & Ratio 1:2 & 89.36 $\pm$ 1.15 & 89.73 $\pm$ 1.20 \\
Both & Ratio 2:1 & 88.08 $\pm$ 1.39 & 88.83 $\pm$ 1.13 \\
Both & Dynamic Threshold & 88.64 $\pm$ 2.21 & 89.48 $\pm$ 1.92 \\
Both & Consistent Labeling & 88.56 $\pm$ 1.51 & 88.80 $\pm$ 1.69 \\
Both & At Least One Match & 89.04 $\pm$ 1.40 & 89.51 $\pm$ 1.25 \\
H-only & Ratio 1:2 & 88.56 $\pm$ 1.66 & 89.25 $\pm$ 1.40 \\
H-only & Ratio 2:1 & 88.96 $\pm$ 0.60 & 89.46 $\pm$ 0.56 \\
H-only & Dynamic Threshold & 89.84 $\pm$ 1.56 & 90.27 $\pm$ 1.42 \\
H-only & Consistent Labeling & 88.08 $\pm$ 0.95 & 88.52 $\pm$ 1.01 \\
H-only & At Least One Match & 88.24 $\pm$ 1.45 & 88.62 $\pm$ 1.47 \\
L-only & Ratio 1:2 & 88.80 $\pm$ 1.16 & 89.25 $\pm$ 1.21 \\
\textbf{L-only} & \textbf{Ratio 2:1} & \textbf{89.92 $\pm$ 0.76} & \textbf{90.40 $\pm$ 0.64} \\
L-only & Dynamic Threshold & 89.68 $\pm$ 0.91 & 90.01 $\pm$ 0.93 \\
L-only & Consistent Labeling & 88.16 $\pm$ 0.73 & 88.78 $\pm$ 0.75 \\
L-only & At Least One Match & 87.52 $\pm$ 1.42 & 88.05 $\pm$ 1.31 \\
\bottomrule
\end{tabular}
\end{table}

\begin{table}[ht]
\centering
\footnotesize  
\caption{Comparison with Competing Methods on Our Dataset}  
\label{2}
\begin{tabular}{lcc}
\toprule
\textbf{Model} & \textbf{BA(\%)} $\uparrow$ & \textbf{F1(\%)} $\uparrow$ \\
\midrule
DGNet \cite{ji2023gradient}& 78.80 & 76.10 \\
SINet-V2 \cite{fan2021concealed}& 72.40 & 69.10 \\
ResNet-18 \cite{he2016deep}& 86.00 $\pm$ 0.88 & 86.44 $\pm$ 0.72 \\
ResNeXt-50 \cite{xie2017aggregated}& 86.56 $\pm$ 1.51 & 87.16 $\pm$ 1.13 \\
DenseNet121 \cite{huang2017densely}& 87.76 $\pm$ 0.60 & 88.12 $\pm$ 0.84 \\
EfficientNetB0 \cite{tan2019efficientnet}& 84.40 $\pm$ 1.52 & 84.44 $\pm$ 1.47 \\
SwinT \cite{Liu_2021_ICCV}& 88.00 $\pm$ 1.77 & 88.83 $\pm$ 1.54 \\
ViT-B16 \cite{dosovitskiy2020image}& 86.40 $\pm$ 1.13 & 86.56 $\pm$ 1.25 \\
\textbf{SwinT + Training Policy} & \textbf{89.92 $\pm$ 0.76} & \textbf{90.40 $\pm$ 0.64} \\
\textbf{SwinT + Training Policy + RSVP(Mean)} & \textbf{92.56 $\pm$ 1.12} & \textbf{92.49 $\pm$ 1.17} \\
\textbf{SwinT + Training Policy + RSVP(Best)} & \textbf{95.60$\pm$0.15} & \textbf{95.49$\pm$0.16}\\
\bottomrule
\end{tabular}
\end{table}

\subsection{Comparison with Existing Methods and Ablation Study}
We evaluated our approach against two categories of existing methods, as summarized in Table~\ref{2}. The first category includes camouflaged object segmentation models, represented by DGNet and SINet-V2. These models generate completely black output maps for background images, requiring a dynamic threshold (optimized on the training set) to analyze the proportion of white pixels in the Ground Truth image and detect camouflaged targets. The second category comprises widely used CV models, including CNN-based architectures such as ResNet-18, ResNeXt-50, and DenseNet121; lightweight models such as EfficientNetB0; and transformer-based models, such as the Swin Transformer (SwinT) and Vision Transformer (ViT-B16). 

For these CV models, we perform the following steps: 1) load the models and use pre-trained weights from ImageNet1k; 2) freeze the pre-trained weights; 3) retrieve the number of input features for the classification head; 4) replace the original classification head with a new fully connected layer consisting of four linear transformation layers, which map to 256, 32, and 8 dimensions, and finally output two classes; 5) Unfreeze the parameters of the new classification head. The key findings of the experiments include the following.

\begin{itemize}
\item Performance of COD Models: COD models, such as DGNet and SINet-V2, achieved approximately 75\% balanced accuracy (BA) in detecting the presence or absence of camouflaged targets. Although effective in edge detection and segmentation, their reliance on local pixel-level features limits their ability to perform well in tasks that require a broader contextual understanding.

\item Performance of CV Models: Among the CV models tested, SwinT exhibited the best performance in our data set and was selected as the backbone model for our approach.

\item Impact of Training Policy: The confidence-based training policy applied to SwinT improved BA by 1.92\% and the F1 score by 1.57\%, demonstrating the ability of the policy to enhance model robustness and precision.

\item RSVP Integration: Combining RSVP with the training policy resulted in additional improvements of 2.64\% in BA and 2.09\% in F1 score on average. The best participant achieved a remarkable balanced accuracy of 95.60\% and an F1 score of 95.49\%, setting a new state-of-the-art in this evaluation.
\end{itemize}
\subsection{Results for RSVP-based BCIs}
To determine the optimal EEG model, we compared several leading EEG analysis methods. EEGNet, designed specifically for EEG signal classification, demonstrated outstanding performance. PLNet leverages phase-locking features of Event-Related Potentials (ERPs) for spatio-temporal feature extraction, excelling in single-session RSVP EEG classification. PPNN, a pyramid-structured parallel neural network, captures multiscale spatio-temporal features, improving classification. EEG-Inception adapts computer vision concepts to ERP detection, improving classification accuracy for ERP-based BCIs. LMDA-Net (LMDA) combines channel and depth attention modules to improve classification by integrating multidimensional features. EEG-Conformer (Conformer) combines convolutional networks and transformers, capturing both local and long-range dependencies, thus improving classification performance.

Table~\ref{3} shows the BA and F1 scores for each model across all participants. As seen in Table~\ref{3}, EEGNet performed the best for six participants, while Conformer excelled for two. Table~\ref{4} provides the mean performance statistics for each model. Among all models, EEGNet achieved the highest performance in the RSVP-based BCIs task, with a BA of 76.76\% and the highest classification precision. Although PLNet, PPNN and EEG-Inception had a lower accuracy, LMDA and Conformer showed better F1 scores but did not outperform EEGNet in accuracy. Based on these results, we selected EEGNet as the optimal EEG model for our task.

\begin{table}[H]
\centering
\caption{Comparison of Classification Accuracy Across Different EEG Models for Various Participants}  
\footnotesize  
\label{3}
\setlength{\tabcolsep}{1mm}  
\begin{tabular}{ccccccc}  
\toprule
 & \multicolumn{2}{c}{Subject 1} & \multicolumn{2}{c}{Subject 2} & \multicolumn{2}{c}{Subject 3} \\
\cmidrule{2-3}\cmidrule{4-5}\cmidrule{6-7}
& BA(\%) $\uparrow$ & $F_1$(\%) $\uparrow$ & BA(\%) $\uparrow$ & $F_1$(\%) $\uparrow$ & BA(\%) $\uparrow$ & $F_1$(\%) $\uparrow$  \\ \midrule
EEGNet & 63.66$\pm$1.33 & 58.26$\pm$1.10 & 85.41$\pm$0.77 & 80.25$\pm$1.31 & 77.77$\pm$1.14 & 69.12$\pm$1.38 \\
PLNet & 60.63$\pm$2.05 & 56.62$\pm$1.61 & 84.02$\pm$1.24 & 78.31$\pm$3.51 & 74.79$\pm$1.55 & 68.80$\pm$2.08 \\
PPNN & 55.54$\pm$1.83 & 54.68$\pm$1.22 & 78.55$\pm$1.44 & 79.58$\pm$1.14 & 71.19$\pm$1.82 & 71.09$\pm$1.61 \\
EEGInception & 59.25$\pm$4.05 & 56.63$\pm$5.76 & 72.61$\pm$6.54 & 74.68$\pm$8.01 & 73.29$\pm$6.13 & 72.56$\pm$3.34 \\
LMDA & 58.41$\pm$1.99 & 59.03$\pm$2.18 & 83.89$\pm$2.50 & 84.48$\pm$1.86 & 69.64$\pm$1.05 & 71.38$\pm$1.20 \\
Conformer & 67.63$\pm$1.91 & 61.26$\pm$1.45 & 88.82$\pm$1.11 & 85.60$\pm$1.76 & 74.67$\pm$2.32 & 71.10$\pm$1.84 \\
\\
\toprule
 & \multicolumn{2}{c}{Subject 4} & \multicolumn{2}{c}{Subject 5} & \multicolumn{2}{c}{Subject 6} \\
\cmidrule{2-3}\cmidrule{4-5}\cmidrule{6-7}
& BA(\%) $\uparrow$ & $F_1$(\%) $\uparrow$ & BA(\%) $\uparrow$ & $F_1$(\%) $\uparrow$ & BA(\%) $\uparrow$ & $F_1$(\%) $\uparrow$ \\ \midrule
EEGNet & 83.17$\pm$1.47 & 75.21$\pm$1.74 & 80.57$\pm$1.62 & 69.59$\pm$1.64 & 79.98$\pm$1.47 & 66.76$\pm$2.18 \\
PLNet & 80.97$\pm$2.14 & 74.72$\pm$1.39 & 78.38$\pm$2.55 & 69.59$\pm$2.20 & 71.96$\pm$2.57 & 64.12$\pm$1.41 \\
PPNN & 77.70$\pm$3.07 & 77.64$\pm$1.65 & 74.02$\pm$1.77 & 71.93$\pm$1.78 & 66.66$\pm$3.04 & 66.32$\pm$2.17 \\
EEGInception & 76.03$\pm$5.85 & 74.61$\pm$4.84 & 72.91$\pm$5.52 & 70.31$\pm$3.91 & 76.40$\pm$6.20 & 66.89$\pm$9.17 \\
LMDA & 74.38$\pm$1.92 & 74.85$\pm$1.68 & 71.84$\pm$2.57 & 71.63$\pm$1.85 & 72.16$\pm$2.07 & 70.80$\pm$1.82 \\
Conformer & 79.09$\pm$1.59 & 77.04$\pm$0.98 & 76.97$\pm$1.92 & 71.85$\pm$1.48 & 75.35$\pm$3.20 & 69.11$\pm$1.96 \\
\\
\toprule
 & \multicolumn{2}{c}{Subject 7} & \multicolumn{2}{c}{Subject 8} \\
\cmidrule{2-3}\cmidrule{4-5}
& BA(\%) $\uparrow$ & $F_1$(\%) $\uparrow$ & BA(\%) $\uparrow$ & $F_1$(\%) $\uparrow$ \\ \cmidrule{1-5}
EEGNet & 70.56$\pm$1.31 & 63.13$\pm$1.31 & 72.98$\pm$1.62 & 65.21$\pm$1.38 \\
PLNet & 62.81$\pm$2.79 & 58.60$\pm$1.06 & 67.38$\pm$2.41 & 62.53$\pm$1.13 \\
PPNN & 62.19$\pm$2.47 & 61.98$\pm$2.08 & 61.89$\pm$1.95 & 61.37$\pm$1.52 \\
EEGInception & 64.01$\pm$5.22 & 61.36$\pm$3.44 & 69.94$\pm$3.79 & 64.48$\pm$7.97 \\
LMDA & 59.63$\pm$1.99 & 60.68$\pm$2.17 & 67.40$\pm$1.78 & 66.53$\pm$1.87 \\
Conformer & 66.58$\pm$2.08 & 63.36$\pm$1.27 & 69.59$\pm$1.31 & 65.61$\pm$1.26 \\
\bottomrule
\end{tabular}
\end{table}

\begin{table}[ht]
\caption{Summary of the Mean Performance Statistics for Each EEG Classification Model}  
\footnotesize  
\label{4}
\centering
\begin{tabular}{lcc}
\toprule
\textbf{EEG Model} & \textbf{BA(\%)} $\uparrow$ & \textbf{F1(\%)} $\uparrow$ \\
\midrule
\textbf{EEGNet} \cite{lawhern2018eegnet}& \textbf{76.76$\pm$6.92} & \textbf{68.44$\pm$6.64} \\
PLNet \cite{zang2021deep}& 72.61$\pm$8.26 & 66.66$\pm$7.37 \\
PPNN \cite{li2021phase}& 68.46$\pm$8.06 & 68.07$\pm$8.20 \\
EEGInception\cite{santamaria2020eeg} & 70.56$\pm$7.76 & 67.69$\pm$8.60 \\
LMDA \cite{miao2023lmda}& 69.67$\pm$7.93 & 69.92$\pm$7.80 \\
\textbf{Conformer} \cite{song2022eeg}& \textbf{74.83$\pm$7.07} & \textbf{70.62$\pm$7.55} \\
\bottomrule
\end{tabular}
\end{table}

\subsection{Results for Human-machine Collaboration}

For human-machine collaboration, we selected the highest proportion of samples with the highest uncertainty and replaced their CV-predicted labels with EEG-predicted labels. The results for different proportions of uncertain samples are shown in Table~\ref{5}. The model performed optimally when the correction proportion was set to 20\%. A lower proportion did not fully capture the advantages of human input, while a higher proportion reduced the contribution of the CV model and significantly increased manual effort. Fortunately, the 20\% correction ratio provided an effective balance, allowing human input without excessive stress. Importantly, for all proportions tested, the model outperformed the baseline CV model (SwinT with our training policy), demonstrating the general benefits of human-machine collaboration.

\begin{table*}[ht]
\caption{Test Set Results Comparing the Replacement of Low-Confidence Image Predictions with EEG-Predicted Labels at Different Uncertainty Ratios} 
\footnotesize  
\label{5}
\setlength{\tabcolsep}{1mm}  
\begin{tabular*}{\textwidth}{@{\extracolsep\fill}lcccc}
\toprule
 & \multicolumn{2}{@{}c@{}}{10\%} & \multicolumn{2}{@{}c@{}}{20\%}\\
\cmidrule{2-3}\cmidrule{4-5}
 & BA(\%) & $F_1$(\%) & BA(\%) & $F_1$(\%) \\ \midrule
Subject1 & 91.52$\pm$0.29 & 91.63$\pm$0.28 & 91.52$\pm$0.13 & 91.42$\pm$0.17\\
Subject2 & 92.64$\pm$0.29 & 92.73$\pm$0.28 & 94.64$\pm$0.24 & 94.60$\pm$0.24\\
Subject3 & 91.36$\pm$0.27 & 91.39$\pm$0.27 & 92.00$\pm$0.38 & 91.78$\pm$0.41\\
Subject4 & 91.52$\pm$0.21 & 91.71$\pm$0.21 & 92.72$\pm$0.23 & 92.86$\pm$0.23\\
Subject5 & 92.48$\pm$0.13 & 92.66$\pm$0.13 & 94.24$\pm$0.21 & 94.30$\pm$0.22\\
Subject6 & 92.88$\pm$0.07 & 93.01$\pm$0.07 & 93.28$\pm$0.18 & 93.31$\pm$0.17\\
Subject7 & 91.76$\pm$0.27 & 91.88$\pm$0.28 & 90.96$\pm$0.24 & 90.83$\pm$0.25\\
Subject8 & 92.56$\pm$0.33 & 92.72$\pm$0.32 & 93.04$\pm$0.33 & 93.00$\pm$0.33\\
Mean & 92.09$\pm$0.62 & 92.22$\pm$0.63 & \textbf{92.80$\pm$1.23} & \textbf{92.76$\pm$1.28}\\
\\
\toprule
 & \multicolumn{2}{@{}c@{}}{30\%} & \multicolumn{2}{@{}c@{}}{40\%} \\
 \cmidrule{2-3}\cmidrule{4-5}
 & BA(\%) & $F_1$(\%) & BA(\%) & $F_1$(\%)\\ \midrule
Subject1 & 90.48$\pm$0.13 & 90.24$\pm$0.13 & 88.16$\pm$0.09 & 87.54$\pm$0.10\\
Subject2 & 95.60$\pm$0.15 & 95.49$\pm$0.16 & 94.40$\pm$0.11 & 94.12$\pm$0.13\\
Subject3 & 90.96$\pm$0.21 & 90.41$\pm$0.25 & 89.92$\pm$0.13 & 89.18$\pm$0.14\\
Subject4 & 93.36$\pm$0.09 & 93.42$\pm$0.09 & 93.20$\pm$0.16 & 93.18$\pm$0.16\\
Subject5 & 93.12$\pm$0.13 & 93.08$\pm$0.12 & 91.20$\pm$0.11 & 91.00$\pm$0.11\\
Subject6 & 92.48$\pm$0.18 & 92.40$\pm$0.20 & 88.16$\pm$0.09 & 90.32$\pm$0.13\\
Subject7 & 90.00$\pm$0.34 & 89.49$\pm$0.36 & 88.08$\pm$0.13 & 87.21$\pm$0.15\\
Subject8 & 92.00$\pm$0.11 & 91.66$\pm$0.14 & 90.16$\pm$0.18 & 89.53$\pm$0.22\\
Mean & \textbf{92.25$\pm$1.72} & \textbf{92.02$\pm$1.89} & 90.41$\pm$2.27 & 90.26$\pm$2.31\\
\bottomrule
\end{tabular*}
\end{table*}

\section{Discussion}

%
\subsection{Application Scenarios}
We propose that COD can be divided into two subtasks: identification and location. Our focus is on the binary classification task of determining whether a camouflaged object is present. In scenarios where there is no prior knowledge about the presence of a camouflaged object in an image, a "classify-then-segment" approach aligns better with practical application requirements. Additionally, since the computational cost and runtime of our classification model are significantly lower than those of a segmentation model, this approach also helps conserve computational resources. Furthermore, the integration of our identification module with other location models has the potential to be refined more. For example, an interaction between the location model uncertainty map and the identification module classification uncertainty could be designed collaboratively to enhance the detection performance.

\subsection{Uncertainty Sources}
Uncertainty in machine learning can generally be categorized as aleatoric or epistemic, each arising from different sources. In particular, the uncertainties involved in the identification of camouflaged objects and location detection are different. For camouflaged object identification, aleatoric uncertainty is particularly high when the camouflaged objects are indistinct and difficult to distinguish, such as animals in a forest blending seamlessly with their surroundings, like branches or foliage. In these scenarios, the inherent ambiguity in the environment and the multiple possibilities reflected in the training data contribute to this type of uncertainty. In contrast, epistemic uncertainty stems from a lack of knowledge of unseen or unfamiliar data. For example, it arises when the camouflaged scenes or objects differ entirely from those in the training dataset, such as new backgrounds or novel camouflage techniques. In this work, we use the term "predictive uncertainty" to refer to the overall uncertainty in a given situation, encompassing both aleatoric and epistemic components.

\subsection{Training process}
Table 6 illustrates how the precision of samples within different confidence intervals changes over training epochs in the validation set. The uncertainty is ranked in ascending order, where higher percentages indicate lower confidence levels in the CV model's predictions. The confusion matrix (CM) is represented in the format
$\begin{bmatrix}
TP & FN \\ 
FP & TN 
\end{bmatrix}$, where the true positives (TP), false negatives (FN), false positives (FP), and true negatives (TN) are laid out in matrix form.

A clear trend emerges: the confidence level of the CV model is inversely correlated with prediction accuracy. This highlights the effectiveness and robustness of our multiview uncertainty measurement approach. Moreover, as training progresses, the accuracy of the top 80\% most confident samples improves, while the accuracy of the bottom 20\% least confident samples decreases significantly. A similar pattern is observed on the test set. This decline in accuracy for low-confidence samples underpins the foundation for human-machine collaboration, as it identifies cases where the CV model lacks confidence and could benefit from human intervention.

\begin{table*}[ht]
\centering
\footnotesize
\caption{Accuracy Trends of Samples Across Confidence Intervals During Training on the Validation Set. Uncertainty Is Ranked in Ascending Order, with Higher Percentages Indicating Lower Model Confidence. The Confusion Matrix (CM) Is Represented as $\begin{bmatrix} TP & FN \\ FP & TN \end{bmatrix}$.}
\begin{tabular}{>{\centering\arraybackslash}p{0.5cm} 
                >{\centering\arraybackslash}p{0.4cm} 
                >{\centering\arraybackslash}p{0.4cm} 
                >{\centering\arraybackslash}p{1.2cm}  
                >{\centering\arraybackslash}p{0.4cm} 
                >{\centering\arraybackslash}p{0.4cm} 
                >{\centering\arraybackslash}p{1.2cm}
                >{\centering\arraybackslash}p{0.4cm} 
                >{\centering\arraybackslash}p{0.4cm} 
                >{\centering\arraybackslash}p{1.2cm} }
\toprule
& \multicolumn{3}{c}{0-20\%} & \multicolumn{3}{c}{20-40\%} & \multicolumn{3}{c}{40-60\%}\\
\cmidrule{2-4} \cmidrule{5-7} \cmidrule{8-10}
Epoch & BA & F1 & \multicolumn{1}{c}{CM} & BA & F1 & CM & BA & F1 & \multicolumn{1}{c}{CM}\\ \midrule
\multirow{2}{*}{0} & \multirow{2}{*}{97.2} & \multirow{2}{*}{98.1} & \multirow{2}{*}{$\left[\begin{matrix} 27 & 0 \\ 1 & 17 \end{matrix}\right]$} & \multirow{2}{*}{92.8} & \multirow{2}{*}{94.1} & \multirow{2}{*}{$\left[\begin{matrix} 24 & 0 \\ 3 & 18 \end{matrix}\right]$}  & \multirow{2}{*}{90.4} & \multirow{2}{*}{92.3} & \multirow{2}{*}{$\left[\begin{matrix} 24 & 0 \\ 4 & 17 \end{matrix}\right]$}\\
&&&&&&&&&\\
\multirow{2}{*}{10} & \multirow{2}{*}{94.7} & \multirow{2}{*}{96.2} & \multirow{2}{*}{$\left[\begin{matrix} 26 & 0 \\ 2 & 17 \end{matrix}\right]$} & \multirow{2}{*}{96.4} & \multirow{2}{*}{98.4} & \multirow{2}{*}{$\left[\begin{matrix} 31 & 0 \\ 1 & 13 \end{matrix}\right]$}  & \multirow{2}{*}{95.2} & \multirow{2}{*}{96.0} & \multirow{2}{*}{$\left[\begin{matrix} 24 & 0 \\ 2 & 19 \end{matrix}\right]$}\\
&&&&&&&&&\\
\multirow{2}{*}{20} & \multirow{2}{*}{100} & \multirow{2}{*}{100} & \multirow{2}{*}{$\left[\begin{matrix} 18 & 0 \\ 0 & 27 \end{matrix}\right]$} & \multirow{2}{*}{92.1} & \multirow{2}{*}{94.5} & \multirow{2}{*}{$\left[\begin{matrix} 26 & 0 \\ 3 & 16 \end{matrix}\right]$}  & \multirow{2}{*}{97.5} & \multirow{2}{*}{98.0} & \multirow{2}{*}{$\left[\begin{matrix} 25 & 0 \\ 1 & 19 \end{matrix}\right]$}\\
&&&&&&&&&\\
\multirow{2}{*}{31} & \multirow{2}{*}{100} & \multirow{2}{*}{100} & \multirow{2}{*}{$\left[\begin{matrix} 17 & 0 \\ 0 & 27 \end{matrix}\right]$} & \multirow{2}{*}{97.3} & \multirow{2}{*}{98.1} & \multirow{2}{*}{$\left[\begin{matrix} 26 & 0 \\ 1 & 18 \end{matrix}\right]$}  & \multirow{2}{*}{92.1} & \multirow{2}{*}{94.5} & \multirow{2}{*}{$\left[\begin{matrix} 26 & 0 \\ 3 & 16 \end{matrix}\right]$}\\
&&&&&&&&&\\
\\
\toprule
 & \multicolumn{3}{c}{60-80\%} & \multicolumn{3}{c}{80-100\%} \\
\cmidrule{2-4} \cmidrule{5-7}
Epoch & BA & F1 & \multicolumn{1}{c}{CM} & BA & F1 & \multicolumn{1}{c}{CM}\\ \midrule
\multirow{2}{*}{0} & \multirow{2}{*}{86.0} & \multirow{2}{*}{88.4} & \multirow{2}{*}{$\left[\begin{matrix} 23 & 1 \\ 5 & 16 \end{matrix}\right]$} & \multirow{2}{*}{70.5} & \multirow{2}{*}{59.2} & \multirow{2}{*}{$\left[\begin{matrix} 8 & 6 \\ 5 & 26 \end{matrix}\right]$}\\
&&&&&&\\
\multirow{2}{*}{10} & \multirow{2}{*}{89.9} & \multirow{2}{*}{86.4} & \multirow{2}{*}{$\left[\begin{matrix} 16 & 1 \\ 4 & 24 \end{matrix}\right]$} & \multirow{2}{*}{63.3} & \multirow{2}{*}{51.6} & \multirow{2}{*}{$\left[\begin{matrix} 8 & 7 \\ 8 & 22 \end{matrix}\right]$}\\
&&&&&&\\
\multirow{2}{*}{20} & \multirow{2}{*}{90.1} & \multirow{2}{*}{92.0} & \multirow{2}{*}{$\left[\begin{matrix} 23 & 0 \\ 4 & 18 \end{matrix}\right]$} & \multirow{2}{*}{66.1} & \multirow{2}{*}{61.5} & \multirow{2}{*}{$\left[\begin{matrix} 12 & 9 \\ 6 & 18 \end{matrix}\right]$} \\
&&&&&&\\
\multirow{2}{*}{31} & \multirow{2}{*}{89.6} & \multirow{2}{*}{89.4} & \multirow{2}{*}{$\left[\begin{matrix} 21 & 0 \\ 5 & 19 \end{matrix}\right]$} & \multirow{2}{*}{53.3} & \multirow{2}{*}{53.3} & \multirow{2}{*}{$\left[\begin{matrix} 12 & 10 \\ 11 & 12 \end{matrix}\right]$} \\
&&&&&& \\
\bottomrule
\end{tabular}
\end{table*}

\subsection{Failure Cases}
\begin{figure}[!htb]
    \centering
    \includegraphics[width=\linewidth]{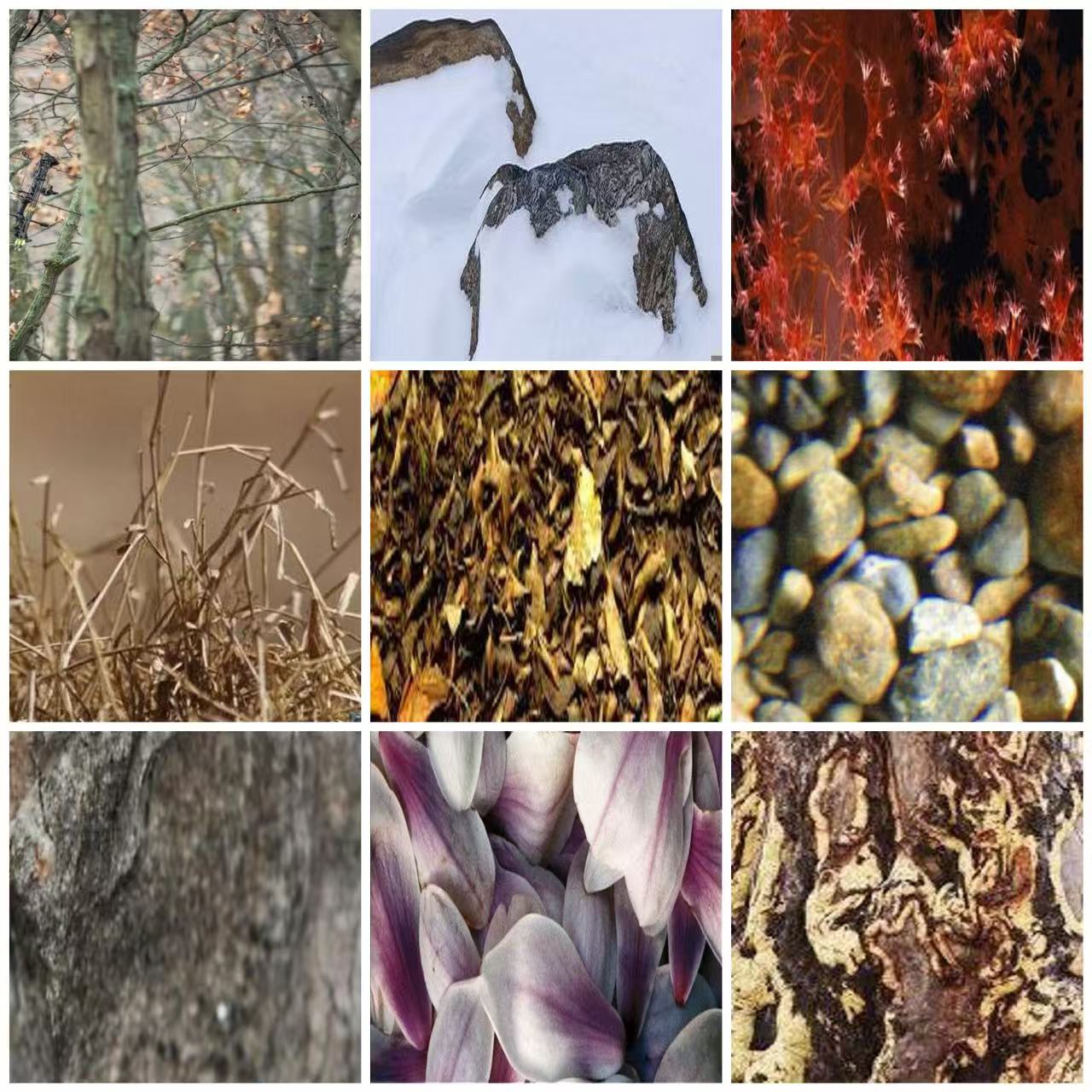} 
    \caption{Examples of CV model failure cases where the CV model incorrectly identified background as containing camouflaged objects.}
    \label{fig:machine_nocam}
\end{figure}
 We observed an interesting phenomenon in the failure cases of the CV model. For instances where there was an actual camouflaged object, but the CV model misclassified it as background, the camouflaged object was also difficult for the human eye to detect. In contrast, for instances where the CV model mistakenly identified the actual background as containing a camouflaged object (examples shown in Figure~\ref{fig:machine_nocam}), humans could easily recognize that no camouflaged object was present. This discrepancy might stem from the CV model's difficulty in distinguishing between salient objects and camouflaged ones. Unlike human vision, which can rely on contextual and semantic cues to identify salient features in an image, the CV model might struggle to capture these subtleties.

\subsection{Challenges and Future work}
We propose a human-machine collaboration framework for COD based on uncertainty estimation. Our method uses a multiview backbone to measure model confidence by analyzing output differences across views, aiding both training and collaboration. Alternative uncertainty estimation methods, such as dropout-based, bootstrap-based, or Gaussian-based approaches, may also be effective. In addition, camouflaged targets are harder to detect than traditional RSVP targets, with lower P300 amplitudes and longer latencies, which may limit human accuracy. Future work may explore EEG responses to camouflaged targets and refine decoding models. We also plan to integrate EEG and eye tracking data to aid in localization, combining them with segmentation models to achieve better human-machine collaboration in both identification and localization.
\section{Conclusion}

This study presents an integration method of RSVP-based BCIs with CV models, where low-confidence samples are redirected to human cognitive input. This approach combines the complementary strengths of humans and machines to tackle challenging detection tasks such as COD. In the CAMO data set, our method outperformed state-of-the-art approaches, with an average improvement of 4.56\% in BA and 3.66\% in the F1 score. For the best-performing participants, the improvements reached 7.6\% in BA and 6.66\% in the F1 score. By allowing humans to focus only on uncertain samples, the method significantly reduces the cognitive load and time required for RSVP tasks. Furthermore, given the variability in the performance of the BCIs due to environmental conditions, user state, and electrode quality, this human-machine collaboration framework enhances the overall robustness and reliability of the system. In summary, this research paves the way for future exploration of neuroscience and human-computer interaction, providing a promising framework for addressing complex detection challenges.

\section{Acknowledgments}
This work was supported by National Natural Science Foundation of China (U20B2074, 62471169), Key Research and Development Project of Zhejiang Province (2023C03026, 2021C03001, 2021C03003), Key Laboratory of Brain Machine Collaborative Intelligence of Zhejiang Province (2020E10010), and supported by Zhejiang Provincial Natural Science Foundation of China (No. LQN25F020013).


\end{document}